\documentclass[10pt,twocolumn,letterpaper]{article}

\usepackage{wacv}
\usepackage{times}
\usepackage{epsfig}
\usepackage{graphicx}
\usepackage{amsmath}
\usepackage{amssymb}
\usepackage{array}
\usepackage{adjustbox}
\usepackage{ifthen}
\usepackage{tabularx}
\usepackage{ragged2e}
\usepackage{float}
\newcommand{\insertcustomtable}[7]{%
\begin{table}[H]
\centering
\begin{tabular}{m{0.20\linewidth}m{0.20\linewidth}m{0.20\linewidth}m{0.20\linewidth}}
\hline
\ifthenelse{\equal{#2}{\empty}}{
}{
& & & \\
\adjustbox{center}{\begin{varwidth}[t]{\linewidth}\centering\footnotesize\texttt{#2}\end{varwidth}} & 
\adjustbox{center}{\begin{varwidth}[t]{\linewidth}\centering\footnotesize\texttt{#3}\end{varwidth}} & 
\adjustbox{center}{\begin{varwidth}[t]{\linewidth}\centering\footnotesize\texttt{#4}\end{varwidth}} & 
\adjustbox{center}{\begin{varwidth}[t]{\linewidth}\centering\footnotesize\texttt{#5}\end{varwidth}} \\
& & & \\
}
\ifthenelse{\equal{#7}{double}}{
\adjustbox{valign=m,center}{\includegraphics[width=1\linewidth]{imgs/#1/#1-1.png}} & 
\adjustbox{valign=m,center}{\includegraphics[width=1\linewidth]{imgs/#1/#1-3.png}} & 
\adjustbox{valign=m,center}{\includegraphics[width=1\linewidth]{imgs/#1/#1-5.png}} &
\adjustbox{valign=m,center}{\includegraphics[width=1\linewidth]{imgs/#1/#1-7.png}} \\
\adjustbox{valign=m,center}{\includegraphics[width=1\linewidth]{imgs/#1/#1-2.png}} & 
\adjustbox{valign=m,center}{\includegraphics[width=1\linewidth]{imgs/#1/#1-4.png}} & 
\adjustbox{valign=m,center}{\includegraphics[width=1\linewidth]{imgs/#1/#1-6.png}} &
\adjustbox{valign=m,center}{\includegraphics[width=1\linewidth]{imgs/#1/#1-8.png}} \\
}{
\adjustbox{valign=m,center}{\includegraphics[width=1\linewidth]{imgs/#1/#1-1.png}} & 
\adjustbox{valign=m,center}{\includegraphics[width=1\linewidth]{imgs/#1/#1-2.png}} & 
\adjustbox{valign=m,center}{\includegraphics[width=1\linewidth]{imgs/#1/#1-3.png}} &
\adjustbox{valign=m,center}{\includegraphics[width=1\linewidth]{imgs/#1/#1-4.png}} \\
}

\hline
\end{tabular}
\vspace{0.2cm}
\caption{#6}
\label{tab:#1}
\end{table}
}

\wacvfinalcopy

\usepackage[breaklinks=true, hidelinks=true, bookmarks=false]{hyperref}

\begin{document}

\title{ELODIN: Naming Concepts in Embedding Spaces}

\author{Rodrigo Mello\\
Centro de Informática, UFPE\\
Brazil\\
{\tt\small rvcam@cin.ufpe.br}
\and
Filipe Calegario\\
Centro de Informática, UFPE\\
Brazil\\
{\tt\small fcac@cin.ufpe.br}
\and
Geber Ramalho\\
Centro de Informática, UFPE\\
Brazil\\
{\tt\small glr@cin.ufpe.br}
}

\makeatletter
\let\@oldmaketitle\@maketitle
\renewcommand{\@maketitle}{\@oldmaketitle

\centering
\begin{tabular}{m{0.20\linewidth}m{0.20\linewidth}m{0.20\linewidth}m{0.20\linewidth}}
\hline
& & & \\
\adjustbox{center}{\begin{varwidth}[t]{\linewidth}\centering\texttt{<Lucy> at the beach}\end{varwidth}} & 
\adjustbox{center}{\begin{varwidth}[t]{\linewidth}\centering\texttt{<Lucy> at business meeting}\end{varwidth}} & 
\adjustbox{center}{\begin{varwidth}[t]{\linewidth}\centering\texttt{<Lucy> ready for a party}\end{varwidth}} & 
\adjustbox{center}{\begin{varwidth}[t]{\linewidth}\centering\texttt{<Lucy> petting a dog}\end{varwidth}} \\
 & & & \\
\adjustbox{valign=m,center}{\includegraphics[width=1\linewidth]{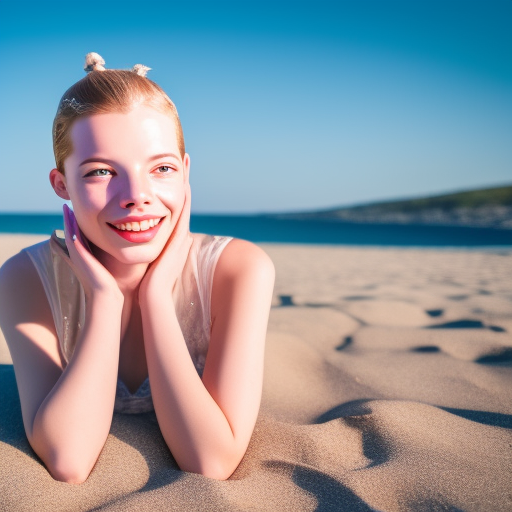}} & 
\adjustbox{valign=m,center}{\includegraphics[width=1\linewidth]{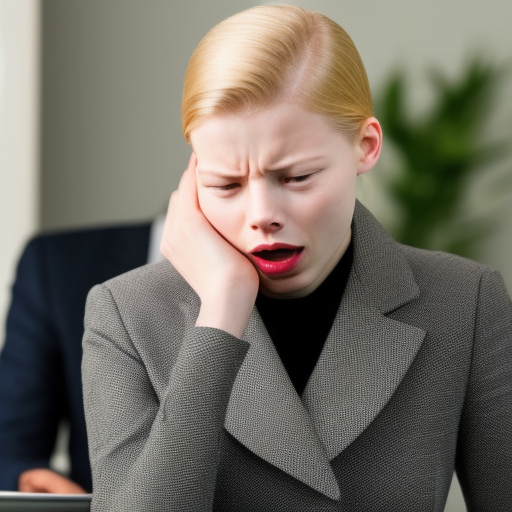}} & 
\adjustbox{valign=m,center}{\includegraphics[width=1\linewidth]{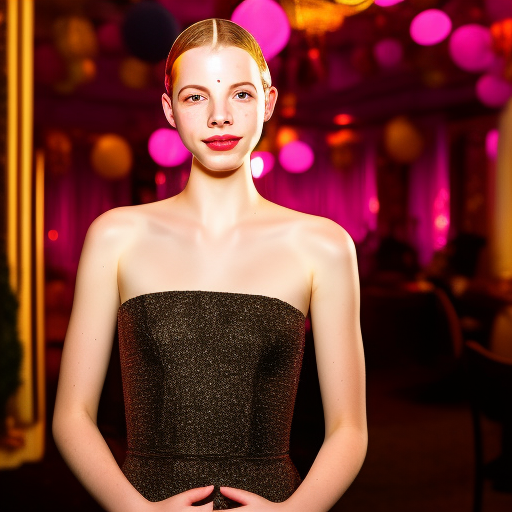}} &
\adjustbox{valign=m,center}{\includegraphics[width=1\linewidth]{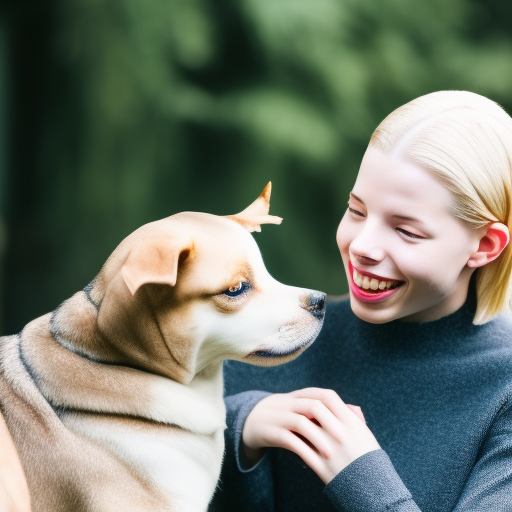}} \\
\adjustbox{valign=m,center}{\includegraphics[width=1\linewidth]{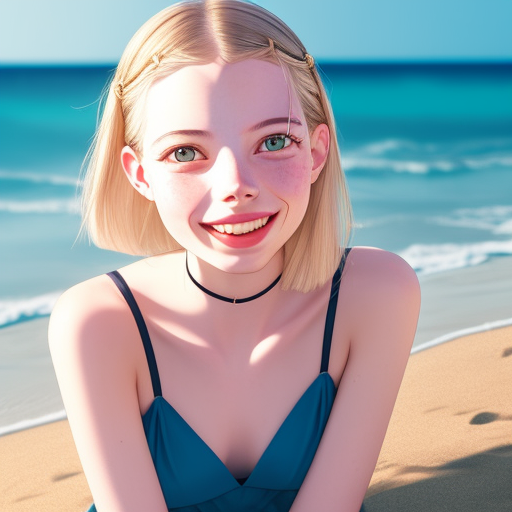}} & 
\adjustbox{valign=m,center}{\includegraphics[width=1\linewidth]{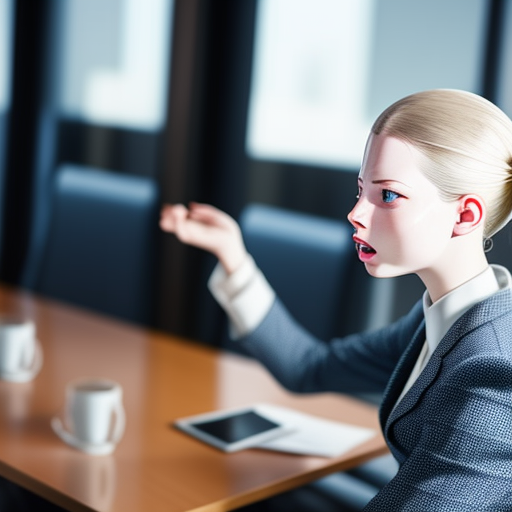}} & 
\adjustbox{valign=m,center}{\includegraphics[width=1\linewidth]{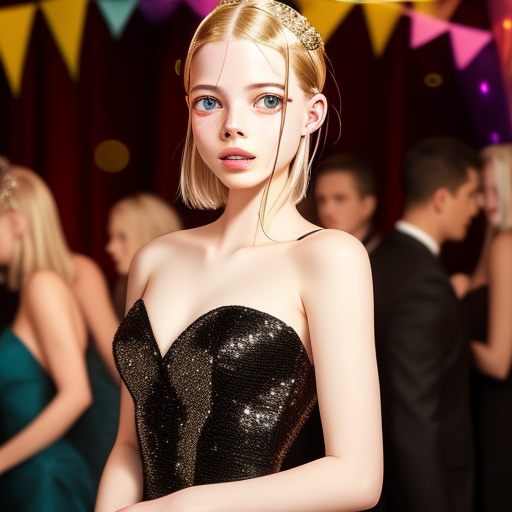}} &
\adjustbox{valign=m,center}{\includegraphics[width=1\linewidth]{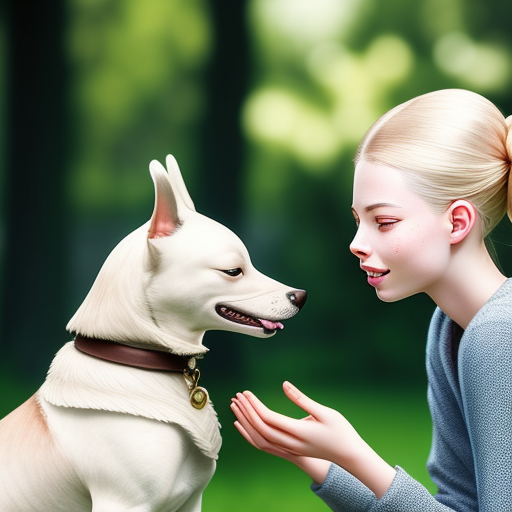}} \\
\hline
\end{tabular}\\
\vspace{0.2cm}
Table 1. The ``named concept'' \texttt{<Lucy>} persists across different prompts and checkpoints (Stable Diffusion 1.4 in the top row and the SD 1.5 variant Dreamshaper in the bottom row). The Dreamshaper model is characterized by a stylized, cartoonish look. The facial appearance of \texttt{<Lucy>} is completely generated, i.e., does not mimic any input face.
  \bigskip}
\makeatother

\maketitle
\addtocounter{table}{1}

\begin{abstract}
\vspace{-1.2cm}
Despite recent advancements, the field of text-to-image synthesis still suffers from lack of fine-grained control. Using only text, it remains challenging to deal with issues such as concept coherence and concept contamination. We propose a method to enhance control by generating specific concepts that can be reused throughout multiple images, effectively expanding natural language with new words that can be combined much like a painter`s palette. Unlike previous contributions, our method does not replicate visuals from input data. In some cases, it can generate concepts through text alone. We perform a set of comparisons that finds our method to be a significant improvement over text-only prompts.
\end{abstract}


\section{Introduction}

The field of generative AI has grown significantly in the past years, experiencing one revolution after the other and achieving remarkable feats. Nevertheless, there are still key challenges to be overcome, including limitations related to fine-grained control such as the ones tackled by many recent editing works \cite{Brooks2022InstructPix2PixLT, Kawar2022ImagicTR, Yang2022PaintBE} and inversion works \cite{TextualInversion_Gal2022, DreamBooth_Ruiz2022, CustomDiffusion_kumari2022}, which look into existing images (generated or otherwise) as a way of creating with precision.

In this context, we aim to provide a way of persisting precise visual concepts in a zero-shot fashion, that is, without relying on pre-existing images. Our solution involves searching certain vectors in the embedding space of encoders, generating special ``named concepts'' that expand natural language vocabulary much like new words. We call this process \emph{concept naming} and our method, ELODIN\footnote{Elodin is a fictional professor from a popular fantasy book (The Name of The Wind) who knows the hidden names of things.}. Our tests have found ELODIN to be successful in that goal. To support our claims, we provide a clear set of side-by-side qualitative comparisons, as well as quantitative measures based on face similarity.


\insertcustomtable{fig2}
{<Lucy> wearing a <galaxy tunic> at the beach}
{<Lucy> in a <fire armor> at a business meeting}
{<Lucy> wearing a dress made of <cloth> for a party}
{<Lucy> wearing red petting a <yellow bird>}
{The same method applied to the task of persisting and combining other concepts. Notice how the elements do not disturb one another (e.g. the colors of the named concept \texttt{<yellow bird>} and the color of the red dress). Notice also how the concepts are also kept similar throughout multiple images and checkpoints, including the top photorealistic (SD 1.4) and the bottom stylized (Dreamshaper) models.}
{double}

\section{Problem Characterization}

Prompting an image synthesis model with natural language descriptions has a number of advantages over the more traditional method of manually defining each pixel's colors (by e.g. brush strokes), as the popularity of recent initiatives such as Stable Diffusion \cite{LDM_Rombach2021HighResolutionIS}, can attest. However, it is not without some disadvantages; notably, the lack of fine-grained control.

We focus on two issues that exemplify this lack of control. The first one, which we refer to as the \textit{concept coherence} issue, speaks to the difficulty of expressing the same visual concept over multiple runs. For instance, an image for a person's face can be easily crafted by an SD user by providing a prompt such as ``a face of a person''. In some cases, however, there is a need to produce more images of that \emph{same person} in different settings (such as at the beach, in a business meeting, or, simply, in another pose). Unfortunately, in these cases, it is challenging to guarantee that the faces in the newly generated images will look alike, even using a long and precise input prompt, as illustrated in Table \ref{tab:fig3}. In this example, although every generated face meets the description, they do not appear as the same person, even using the same seed for each column.

More generally, one may wish to maintain the coherence of any particular visual concept, be it the appearance of a person, object, scene, texture, palette or artistic style, between multiple runs from the same natural language prompt. It is not obvious how to accomplish that, since there is no precise natural language description for most visual concepts (e.g. how to precisely describe a specific face, unless it is the face of a famous person who is well represented in the dataset?).

\insertcustomtable{fig3}{}{}{}{}
{Generations from an extended and descriptive prompt (first row: ``the face of a middle-aged brunette woman with blue eyes and a thin nose at the beach`` and second row ``the face of a middle-aged brunette woman with blue eyes and a thin nose at a business meeting''), with negative prompts ``bad artist, bad perspective'' and the same random seed in each column. SD 1.4}
{double}

This issue affects visual storytelling applications \cite{Jeong2023ZeroshotGO}, such as video clips, games, graphic novels, etc. Such a need is not restricted to storytelling, however; it may impact applications such as product design and publicity.

The other issue, which we call ``\textit{concept contamination}'', arises when a certain visual concept interacts with others in an unintended way. The most obvious example is color: a prompt such as ``a yellow hawk amidst white flowers'' often results in some yellow being applied to the flowers, or in a hawk that is not particularly yellow (Table \ref{tab:fig4}). 

\insertcustomtable{fig4}{}{}{}{}
{prompt ``a yellow hawk amidst white flowers''. SD 1.4 model.}
{single}

In Table \ref{tab:fig5}, we show another more subtle instance of the issue``a warrior wears armor made of fire and lava on a frozen mountain peak''. In a setting like this, the mountain peak might be converted into a volcano because of the mention of lava, or simply be on fire.

\insertcustomtable{fig5}{}{}{}{}
{prompt ``a warrior wears armor made of fire and lava on a frozen mountain peak''. SD 1.4 model.}
{single}

Even though our focus revolves around visual concepts, those issues may in principle be generalized to other kinds of concepts. For example, it is also hard to keep concept coherence in writing style throughout text-to-text generation \cite{Ge2022ExtensiblePF}.

\section{Related Works}

Many solutions have been proposed to generally enhance control in text-to-image models. ControlNet \cite{Zhang2023AddingCC}, for instance, achieves shape control via edge maps, line drawings, segmentation maps, etc., while Latent Guidance \cite{Voynov2022SketchGuidedTD} uses sketches to guide the process. Other works extract style, palette, etc. from one image and apply them to another \cite{Huang2023ComposerCA, Aggarwal2023ControlledAC}. There are also popular tools (OpenAI DALL-E 2\footnote{\url{https://labs.openai.com/}}, DreamStudio\footnote{\url{https://dreamstudio.ai/}}, AUTOMATIC1111 webui \cite{Automatic11112022}) with features such as image-to-image and inpainting, which help provide the model with more information than would be possible with just a simple text prompt.

Focusing on the problem of concept coherence, there is a class of notable methods such as Textual inversion \cite{TextualInversion_Gal2022} and Dreambooth \cite{DreamBooth_Ruiz2022}. Through the process of \emph{inversion}, these methods enable the reproduction of the same visual concept across many runs, as long as this concept can be represented by input images. Those methods often achieve high coherence between multiple representations of the same object. A key limitation among these methods is the need to have the concept already expressed in visual form, often in multiple images. For instance, to generate the same face across multiple runs with Dreambooth, it is necessary to have multiple pictures of that same face from slightly different angles as input data. 

For some of these methods, there are also other limitations, such as difficulty in combining multiple inverted visual concepts in the same run and long training times. Some of these other limitations are explored in concurrent works \cite{CustomDiffusion_kumari2022,Gal2023DesigningAE}. Even so, to the best of our knowledge, no such method works in a zero-shot fashion, that is, they all take as input at least one image representing the visual concept.

The contamination issue is also tackled by inversion techniques, albeit in a cumbersome way. In the ``a yellow hawk amidst white flowers'' example, one would need to first generate some pictures of a yellow hawk, which would then be used as input to the inversion method \cite{Jeong2023ZeroshotGO}. The result of the inversion would, at a later time, be combined with the white flowers. For some situations, the generated objects might be different from one another (such as different armors in the prompt ``a warrior wears armor made of fire and lava on a frozen mountain peak'', or different human faces), thus hindering the inversion process.

\section{Proposal}

Considering the previous challenges, as well as both the limitations and accomplishments of current techniques, we propose a process that is centered on the idea of assigning a custom keyword to a particular concept, even if that concept cannot be easily expressed in natural language. An analogy would be a person's proper noun: even though someone's facial features cannot be easily put into words, their name can be used instead (indeed, when prompted with famous people's names, some models can produce their face as resulting image). Therefore, for example, we can name a generated person \textless Lucy\textgreater.

As another example, to produce a yellow hawk amidst white flowers, one might name the visual concept of a specific yellow hawk (calling it ``\textless my\_hawk\textgreater'', for instance) and, then use that word as part of a prompt (``\textless my\_hawk\textgreater{} amidst white flowers''). All the resulting images then would feature that particular yellow hawk and, since there is no direct mention to its color in the prompt, the flowers would be indeed white. The yellow color, in this case, would be inherent to the bird, much as it is inherent to a banana.

\begin{figure}[!ht]
\begin{center}
\includegraphics[width=1\linewidth]{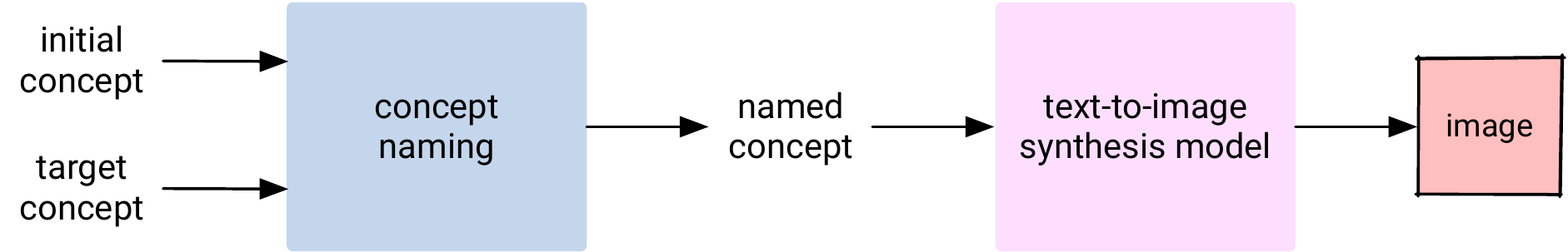}
\end{center}
   \caption{Overall naming process}
\label{fig:elodin_process}
\end{figure}

The whole process is roughly illustrated in Figure \ref{fig:elodin_process}, showing only the inputs and outputs. The boxes will be detailed in Section 6. The role of the \emph{initial concept} (e.g. ``bird'', ``woman'', ``armor'', or ``cloth'') is to provide a starting point for the \emph{concept naming} process. The \emph{target concept} is the specialization of the initial concept one wishes to aim for. That is, the target concept adds qualities to the initial concept, which serves as a course descriptor and to which more qualities are added. Considering the initial concept examples, their respective target concepts could be, for instance: ``a yellow hawk'', ``beautiful blonde woman'', ``armor made of fire and lava'', or ``a very smooth celadon dress. 

When the Elodin process converges, it generates a \emph{named concept} or simply \emph{namecon}, which is a keyword associated to an embedding (an internal neural network representation) reflecting the target concept as precisely as possible. Revisiting our previous examples, one could have respectively the following namecons: \textless my\_hawk\textgreater, \textless Lucy\textgreater, \textless fiery\_armor\textgreater, or \textless my\_cloth\textgreater.

The namecons can be stored to be reused as a vocabulary in future prompts for generating new images (e.g. ``\textless my\_hawk\textgreater{} amidst white flowers'', ``\textless Lucy\textgreater{} in a business meeting'', ``menacing warriors wearing \textless fiery\_armor\textgreater{} atop a frozen mountain peak'', or ``The beautiful woman is wearing garment made of \textless my\_cloth\textgreater''). The namecons can also be freely used together in the same prompt. (``\textless Lucy\textgreater{} wearing a dress made of \textless cloth\textgreater{} while petting \textless my\_hawk\textgreater'').

Differently from previous techniques, the goal of concept naming is not to invert some images into their corresponding inner representations (embeddings), but, rather, to compute representations for new concepts. Therefore, namecons can be created without images, relying only on natural language prompts, or they can be inspired by existing images, without replicating them. 

The benefit of using a natural language prompt to create a namecon instead of the usual method of direct prompting is that using the namecon results in images with more coherence and less contamination (i.e. the control is more precise, since the visual concept is repeatable). The benefit of using an existing image as a target prompt for naming instead of inverting it is not copying precisely the same visual concept of that image. This allows for creating variations that are sufficiently distinguishable from the specific original visual concept, yet retain a broad similarity.

\section{Fundamentals}

To understand the details of the ELODIN method for naming, it is helpful to first consider the following:

\subsection{Text-to-image pipelines}

Recent large-scale text-to-image models, such as the diffusion-based DALL-E 2 \cite{Ramesh2022HierarchicalTI}, Imagen \cite{Imagen_Saharia2022PhotorealisticTD} and Stable Diffusion \cite{LDM_Rombach2021HighResolutionIS} and the GAN-based StyleGAN-T \cite{StyleGANT_Sauer2023} and Parti \cite{Yu2022ScalingAM} are \emph{not} comprised of a single network that translates natural text to image. Instead, in all of those different architectures, the natural language prompt is first input to a language model, such as CLIP \cite{CLIP_Radford2022}, BERT \cite{BERT_Devlin2019}, T5 \cite{Raffel2017OnlineAL}, etc. Those language models then generate an embedding that is used as conditioning for the actual image generation process. Therefore, it is not precise to refer to these mechanisms as text-to-image models, but rather as text-to-image pipelines, since they are not comprised of a single neural network. This distinction becomes very relevant as namecons are an input to the latter, image generation part of the pipeline, instead of being an input to the language model at the start of the pipeline.

\begin{figure}[!ht]
\begin{center}
\includegraphics[width=1\linewidth]{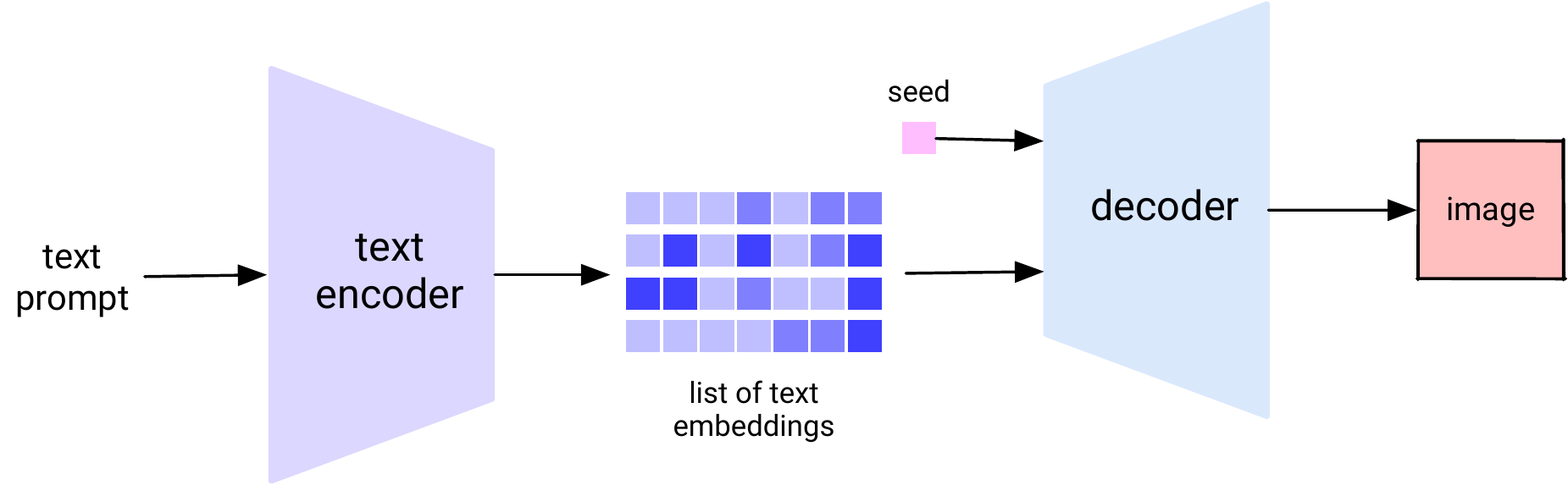}
\end{center}
   \caption{Generic text-to-image pipeline}
\label{fig:generic_process}
\end{figure}

In general terms, as illustrated in Figure \ref{fig:generic_process}, such pipelines can be understood as starting with a \emph{text encoder} component, which translates raw text into a kind of inner representation called \emph{embeddings}. In most implementations, each prompt generates a fixed number of embeddings, no matter the prompt's length. Even though each word roughly corresponds to one of those embeddings, they influence one another because the phrase context is taken into consideration. Each embedding is usually implemented as a number vector of around a thousand dimensions (i.e. a floating-point 1D tensor).

This language component is followed by a \emph{visual decoder} component, which takes the list of embeddings and a random seed (to generate initial noise) and finally outputs an image.

\subsection{Embedding space vs latent space}

\emph{Unconditional} image generation pipelines take noise samples (usually Gaussian) as input and map them to output images, mimicking the probability distribution of the images in the training dataset. The space of possible input samples of noise is called the \emph{latent} space \cite{Image2StyleGAN_Abdal2019}. Many works have used the latent spaces of GANs (especially StyleGAN \cite{StyleGAN2_Karras2019}) to invert and edit images \cite{GANInversion_Xia2021}. More recently, there has been some investigation in Diffusion models' latent spaces as well \cite{Asperti2022ImageEF, Asperti2022ComparingTL}.

\begin{figure*}[!h]
\begin{center}
\includegraphics[width=0.7\linewidth]{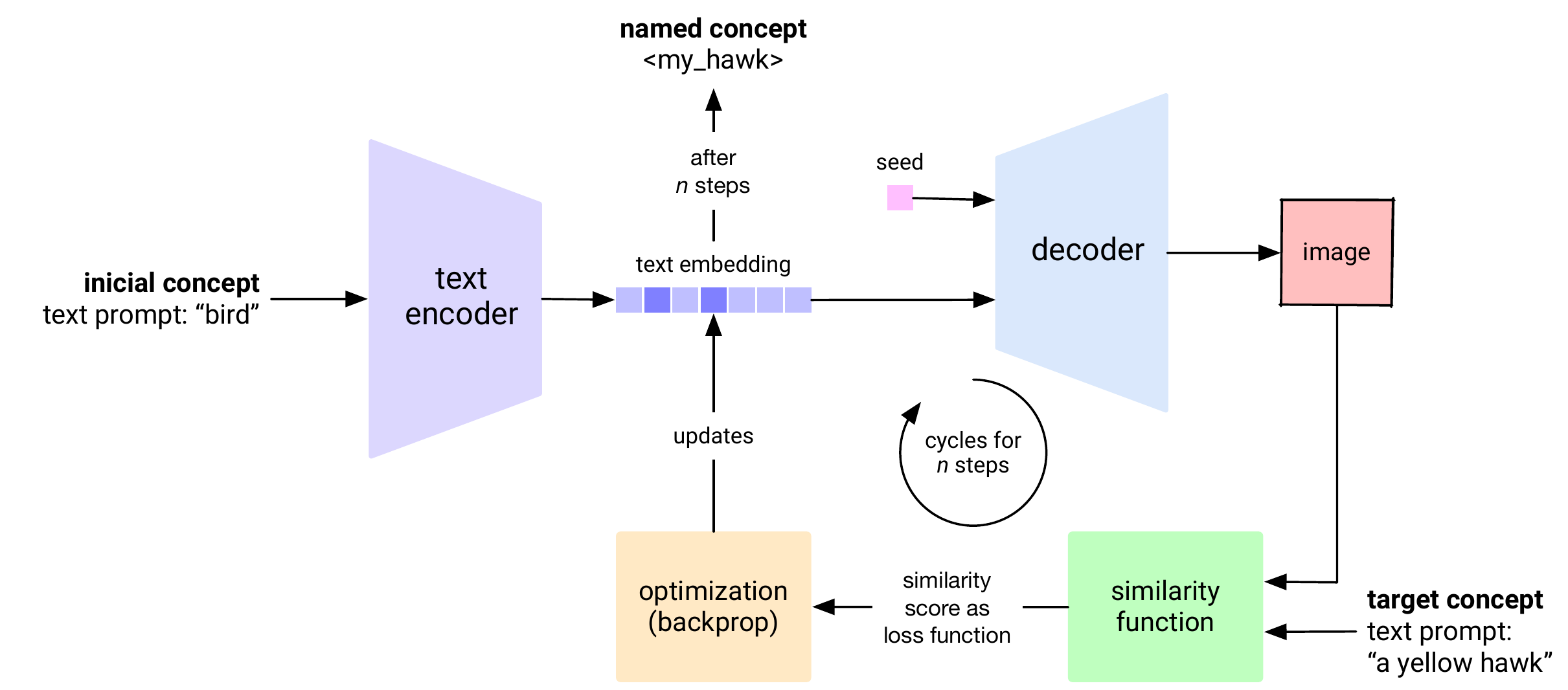}
\end{center}
   \caption{The ELODIN method}
\label{fig:creating_namecon}
\end{figure*}

The concept of latent space is still important for \emph{Conditional} image generation pipelines. However, in this setting, the noise that is sampled from the latent space is no longer the only factor for determining the final output image. The conditioning from the text prompt (that is, the embedding resulting from the language model at the start of the pipeline) is also taken into consideration. For example, the prompt ``a painting of an apple'' will always generate an image that resembles such a painting, independently of the sampled noise from the latent space (though different noise samples will result in different paintings of an apple).

Therefore, we speak of an \emph{embedding space}, comprised of the possible values that can be used as a conditioning for the rest of the text-to-image pipeline. There is a terminology subtlety here: "the possible values the decoder will accept" is not necessarily "the possible values a text encoder will output". Therefore, the embedding space is potentially larger than set of embeddings that can actually be output in practice by any given text-to-image model from natural language descriptions. As hinted at by the aforementioned inversion techniques (which, to the best of our knowledge, follow the seminal work PALAVRA \cite{Cohen2022ThisIM}), the embedding space contains specific notions for particular objects (including fictitious ones) for which there may not be natural language descriptions and that, as such, would not be a result of the processing of a natural language sentence. In concept naming, we are only concerned about sampling from that space, ignoring how it is constructed.

Therefore, our proposed concept naming method can be thought of as a search within the embedding space. In this work, we distinguish between embeddings, which are any vectors in the embedding space, and namecons, which are associations between our computed embeddings and natural language keywords.

\section{Method}

In this section, we go in technical details on our proposal briefly discussed in Section 4.

\subsection{The ELODIN method}

To generate a namecon, we begin by inputting to an encoder a prompt corresponding to the initial concept (as defined in Section 4). Figure \ref{fig:creating_namecon} details the process, taking ``bird' as the initial concept. The encoder generates an embedding (or a list of embeddings), which is then fed to the decoder to produce a batch of images. We compute the similarity between the output images and the ``target concept'' (the target concept would be, for example, ``a yellow hawk''). Finally, we use that similarity score as a loss function to optimize the embedding(s) through backpropagation, generating new images and evaluating new similarity scores at each step until the process reaches a predetermined step count.

Once that count is reached, the user supplies a keyword keyword (e.g. ``my\_hawk'' to be associated with that embedding, creating thus a namecon (e.g \textless my\_hawk\textgreater). The created namecons are stored in a vocabulary of concepts that can be used in the future.

\begin{figure*}[!h]
\begin{center}
\includegraphics[width=0.9\linewidth]{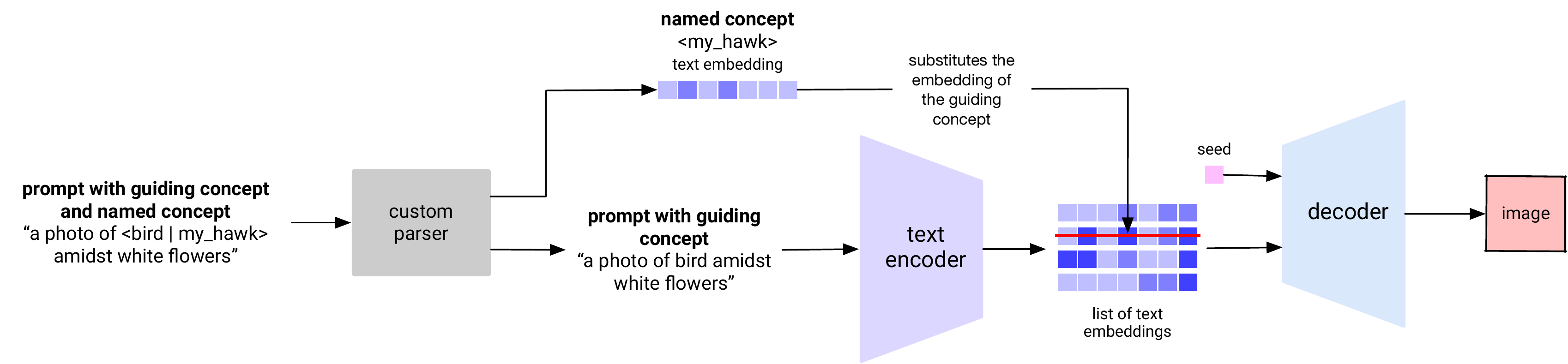}
\end{center}
   \caption{Using the namecon}
\label{fig:using_namecon}
\end{figure*}

Using a different random seed during the optimization process results in a different namecon. In the ``a yellow hawk'' target concept example, ELODIN can produce many specific birds by varying the seed.

Besides the aforementioned similarity loss, we also employ an L2 regularization loss to encourage the final result to have a norm similar to the standard for CLIP embeddings, which is the norm of a multidimensional Gaussian distribution. We found that this helps stabilize the optimization process, that is, without this regularization, the process may result in nonsensical images.

At the end of the naming process, we also directly normalize the resulting namecon to that same Gaussian norm, so it can be better paired up with regular embeddings at inference time. Without this normalization, the namecon may dominate the prompt when combined with regular embeddings, that is, the other embeddings may be ignored.

The similarity scores can be computed in many ways. For instance, one may employ text-image similarity (with a model such as CLIP \cite{CLIP_Radford2022}) to express target concepts through a general text prompt, or, instead, employ face similarity (with Facenet \cite{schroff2015facenet} or others) to express the target as a picture of a face. Therefore, the target concept can be represented in different modalities because it depends on the method we use to compute the similarity score. However, the initial concept is represented as a text prompt because we always use a text encoder to obtain the initial embedding from which the optimization process begins.

In the face identification setting (i.e. using face identification similarity loss instead of text-image similarity loss), the initial concept can be a generic description such as ``blonde woman'' or a proper name of any famous person present in the dataset. The optimization target, in this case, is represented as any front-facing face image. In our method, the optimization does not perfectly reproduce that particular target face. Instead, it converges to a similar --- yet new --- face in the embedding space.

\subsection{Using namecons to generate images}

To use the result of the naming process, i.e. the namecon, we employ a slightly different inference scheme than the usual one proposed by inversion methods. Instead of mapping the embedding to a developer-defined rare word (or ``rare token'' as it is usually called) \cite{TextualInversion_Gal2022}, we allow the user to specify a \emph{guiding concept} for each prompt.

Each prompt follows a syntax as outlined below:

{\small\texttt{\textless a bird \textbar{} my\_hawk\textgreater{} amidst white flowers.}}

In this example, ``a bird'' would be the guiding concept and ``my\_hawk'' would be the namecon's keyword. Our modified text-to-image pipeline includes a parser that ignores the namecon at first, outputting e.g., ``a bird amidst white flowers'' to the language encoder, which computes the sentence's embeddings as normal. Only then do we substitute the embedding(s) corresponding to the guiding concept (``a bird'') for the namecon's embedding and proceed to the visual decoder.

We employ this new inference method because the rare word employed by inversion methods, even if rare, can distort the inference process (since every word influences the embedding of all others). To illustrate this point, consider the results presented in Table \ref{tab:fig10}, which use the namecon of a lava armor, but the guiding concept of a bird.

\insertcustomtable{fig10}{}{}{}{}
{\texttt{<a bird | lava armor>} on a frozen mountain peak}
{single}

For some random seeds, the lava armor takes a bird-like shape. Even though the embeddings for the guiding concept (``a bird'') have been replaced by the namecon, the context of the phrase (i.e. the other embeddings) are influenced by the presence of the guiding concept before the substitution takes place.

The aforementioned rare word method \cite{TextualInversion_Gal2022} is equivalent to specifying a fixed guiding concept that one believes will not influence the results much. In contrast, our method enhances the prompter's control, allowing for an explicit choice of the guiding concept in each prompt.

\section{Experiment Design}

\newcolumntype{C}{>{\centering\arraybackslash}X}

\begin{table*}[ht]
\centering
\begin{tabularx}{\linewidth}{|C|C|C|C|p{0.2\linewidth}|p{0.2\linewidth}|C|}
\hline
\centering\arraybackslash Experiment Name & \centering\arraybackslash Initial Concept & \centering\arraybackslash Target Concept & \centering\arraybackslash Guiding Concept & \centering\arraybackslash Control Prompt & \centering\arraybackslash Proposal Prompt & \centering\arraybackslash Similarity\\ \hline\hline
Bird & 
``bird'' & 
``a yellow hawk'' & 
``a bird'' & 
``a yellow hawk amidst white flowers'' & 
``\textless a bird $|$ my\_hawk\textgreater{} amidst white flowers" & 
text (CLIP)\\ \hline
Armor & 
``fiery armor'' & 
``armor made of fire, lava, and very dark iron'' & 
``armor'' & 
``menacing warriors in armor made of fire, lava, and very dark iron  atop a frozen mountain peak'' & 
``menacing warriors in  \textless armor $|$ my\_armor\textgreater{} atop a frozen mountain peak" & 
text (CLIP)\\ \hline
Cloth & 
``cloth'' & 
``a very smooth celadon dress'' & 
``cloth'' & 
``a photo of a beautiful woman wearing a dress. The beautiful woman is wearing garment made of celadon cloth'' & 
``a photo of a beautiful woman wearing a dress. The beautiful woman is wearing garment made of celadon 
 \textless cloth $|$ my\_cloth \textgreater{}'' & 
text (CLIP)\\ \hline
Lucy & 
the name of a notorious actress & 
a face picture & 
``a woman'' & 
``face of a blonde woman at the beach'' & 
``face of \textless a woman $|$ lucy \textgreater{} at the beach'' & 
face (Facenet)\\ \hline
Stylized Lucy & 
same as the Lucy experiment & 
same as the Lucy experiment & 
same as the Lucy experiment & 
``a blonde woman at a business meeting'' & 
``\textless a woman $|$ lucy \textgreater{} at a business meeting" & 
face (Facenet)\\ \hline
\end{tabularx}
\caption{Parameters for each qualitative3 experiment. We apply the negative prompt ``bad artist, low quality'' in every experiment.}
\end{table*}

To assess how well ELODIN solves the aforementioned concept coherence and concept contamination issues, we perform qualitative side-by-side comparisons between images from an unaltered text-to-image pipeline (``control'') and images from a pipeline modified with our method (``proposal''). We keep all configurations the same (the random seed, negative prompt and model weights), only changing the prompt. In the ``proposal'' setting, we replace each occurrence of the keywords associated with the target prompt by the corresponding namecon. For instance, we compare an image generated using the prompt ``a yellow hawk amidst white flowers'' with another generated using ``\textless a bird $|$ my\_hawk\textgreater{} amidst white flowers''. The prompts were chosen to demonstrate a wide range of visual concepts, such as textures, people, animals and objects.

For naming, we used SD 1.4 for the text setting and a finetuned version of it for the face id setting. Learning rate was set to 4e-2 for text setting and 2e-2 for face id setting, batch size was 1. For image generation, we used the Dreamshaper \cite{DreamShaperCIVITAI} model in the Stylized Lucy experiment, all others employed Stable Diffusion 1.4. Inference was performed through a popular tool \cite{Automatic11112022}.  The diffusion sampler used for naming was DDIM \cite{DDIM_Song2020}. For inference, we used Ancestral Euler \cite{AncestralEuler}.

Even though we do not enlist a large number of crowd workers, we make the generated images fully available in the supplementary material (no cherry-picking). 
We provide a batch of 16 images with corresponding random number generator seeds per experiment.

For the face id setting, we also perform a quantitative analysis based on face similarity. We generate 100 images for both control and proposal. We then use the FaceNet similarity function as a proxy to measure coherence between all pairs inside each group. If our hypothesis is correct, then the average similarity among the proposal group should be higher than among the control group.

The control prompt we use for the quantitative analysis is ``close-up shot of the face of a middle-aged blonde woman with green eyes and thin chin at the park. She is next to a big tree''. The proposal prompt is ``close-up shot of the face of \textless a woman $|$ lucy \textgreater{} at the park. She is next to a big tree``.  For both of those, we employ the negative prompt ``bad artist, low quality``.

\section{Results and Discussion}

In this section, we present and discuss our qualitative and quantitative results.

\subsection{Qualitative Analysis}

Some of the most clear examples from the supplementary are summarized in the following tables. For each experiment, the top row shows the control configuration (no modification), while the bottom row shows the proposal configuration (ours), as discussed in section 7. For each column, the random seed for image generation is the same.

The Bird and Armor experiments highlight the decrease of contamination with ELODIN. In Table \ref{tab:fig11}, notice how the bird keeps its yellow color in the bottom row, white the flowers are much less yellow. In Table \ref{tab:fig12}, notice how the mountain does not catch fire.

\insertcustomtable{fig11}{}{}{}{}
{`Bird' experiment.}
{double}

\insertcustomtable{fig12}{}{}{}{}
{`Armor' experiment.}
{double}

The Lucy and Cloth experiments highlight the increase in coherence. Notice how the person's facial appearance is kept the same in the bottom row (even through different hair colors). Notice also how the texture of the garment's fabric has a more uniform look.

\insertcustomtable{fig13}{}{}{}{}
{`Lucy' experiment}
{double}

\insertcustomtable{fig14}{}{}{}{}
{`Cloth' experiment}
{double}

It is possible to use the same namecon with different decoders as long as they share the same encoder. For example, one may generate images from the same namecon in different finetunings of Stable Diffusion version 1, as demonstrated in Table \ref{tab:fig15}. SD Version 2, however, uses OpenCLIP \cite{ilharco_gabriel_2021_5143773} instead of CLIP as an encoder, so we do not expect namecons to communicate between those versions.

\insertcustomtable{fig15}{}{}{}{}
{`Stylized Lucy' experiment}
{double}

Even though it is possible to generate faces in the text setting, the face id setting produced more natural pictures, while also guaranteeing the generated namecon would represent a face. For instance, in the text setting, the target concept ``a blonde'', may converge to a namecon that consistently results in blonde hair images, instead of images of faces. We hypothesize this may stem from CLIP's low ability to distinguish faces (when compared to a specialized face recognition network)

\subsection{Quantitative Analysis}

As presented in Table \ref{tab:results}, ELODIN scores higher on face similarity than a face generated from a detailed prompt. For comparison, a value of cosine similarity above 0.45 (equivalent to an Euclidean distance below 1.1 \cite{Liang2022MindTG}) would indicate pictures of the same person \cite{schroff2015facenet}.

\begin{table}[t]
\centering
\begin{tabularx}{\linewidth}{|C|C|C|}
\hline
\centering\arraybackslash & \centering\arraybackslash Mean & \centering\arraybackslash Std\\ \hline\hline
Control & 0.43 & 0.14\\ \hline
Proposal & 0.71 & 0.15\\ \hline
\end{tabularx}
\caption{Mean similarity within groups. Higher is better.}
\label{tab:results}
\end{table}


\section{Conclusion}

Motivated by the challenge of fine-grained control in current text-to-image pipelines, our first contribution is to propose the generation of ``\textit{named concepts}'' (\textit{namecons}) in embedding space from minimal input data (such as a text prompt or a single picture of a face). \textit{Naming}, as we call it, is an alternative to the previous two kinds of embedding generation: employing a natural language encoder and directly inverting a set of images. Advantages of naming include the minimal data requirements (sometimes not requiring input images at all), the out-of-the-box capability to combine multiple namecons and the greater diversity of concepts (e.g. many specific faces can be generated from the same input data, changing only generation seed). 

As a second contribution,  we propose a specific method, called ELODIN, to perform naming in the context of visual concepts for image generation. ELODIN amounts to a backpropagation-based search in the embedding space guided by a similarity function (such as text-image similarity or face id similarity) as the loss function. In our experiments, we demonstrate promising results in fine-grained control by applying ELODIN to create and use namecons.

Additional minor contributions are, first, that we deepen the discussion on two issues (concept coherence and concept contamination) in the context of fine-grained control for text-to-image synthesis. Second, that we propose a new inference scheme to use namecons and other custom embeddings (i.e. embeddings not generated by the language encoder), further enhancing control by explicitly providing a \textit{guiding concept}.
 
Regarding future work, we believe our method to be quite general and applicable to other contexts. For instance, it would likely accept more kinds of loss functions beyond text-image and face id similarities, enabling e.g. naming a color shade with a color-similarity loss. Another interesting avenue would be exploring non-visual modalities, such as naming a writing style \cite{Ge2022ExtensiblePF}. 

Additionally, it would be interesting to investigate whether namecons can be used for other tasks beyond generation, such as segmentation or object detection, in the line of previous inversion works \cite{Cohen2022ThisIM}. We also hope to spark more fundamental discussions on the embedding space as a self sufficient entity, separate from the encoder.


{\small
\bibliographystyle{ieee_fullname}
\bibliography{egbib}
}

\end{document}